\documentclass[10pt,twocolumn,letterpaper]{article}

 \usepackage{iccv}              
\usepackage[acronym,nomain,nonumberlist]{glossaries}

\definecolor{iccvblue}{rgb}{0.21,0.49,0.74}
\usepackage[pagebackref,breaklinks,colorlinks,allcolors=iccvblue]{hyperref}
\usepackage{glossaries}
\usepackage{tabularx}
\usepackage[accsupp]{axessibility}  

\usepackage{float}
\glsdisablehyper

\def\paperID{10} 
\def\confName{ICCV}
\def\confYear{2025}

\title{EHWGesture - A dataset for multimodal understanding of clinical gestures}

\author{
Gianluca Amprimo$^{1}$\thanks{Corresponding author.} \quad Alberto Ancilotto$^{2}$ \quad Alessandro Savino$^{1}$ \quad Fabio Quazzolo$^{1}$ \\ \quad Claudia Ferraris $^{3}$ \quad Gabriella Olmo $^{1}$ \quad Elisabetta Farella $^{2}$ \quad Stefano Di Carlo $^{1}$ \\
$^1$Department of Control and Computer Engineering, Politecnico di Torino, Torino, Italy\\
$^2$ Fondazione Bruno Kessler, Trento, Italy \\
$^3$CNR-IEIIT, Torino, Italy \\
{\tt\small {name.surname}@polito.it, \{aancilotto, efarella\}@fbk.eu, \
claudia.ferraris@cnr.it}
}

\newacronym{fps}{fps}{frame per seconds}
\newacronym{AQA}{AQA}{action quality assessment}

\begin{document}
\maketitle
\begin{abstract}

\noindent Hand gesture understanding is essential for several applications in human-computer interaction, including automatic clinical assessment of hand dexterity. While deep learning has advanced static gesture recognition, dynamic gesture understanding remains challenging due to complex spatiotemporal variations. Moreover, existing datasets often lack multimodal and multi-view diversity, precise ground-truth tracking, and an action quality component embedded within gestures.
This paper introduces EHWGesture, a multimodal video dataset for gesture understanding featuring five clinically relevant gestures. It includes over 1,100 recordings ($\sim$6 hours), captured from 25 healthy subjects using two high-resolution RGB-Depth cameras and an event camera. A motion capture system provides precise ground-truth hand landmark tracking, and all devices are spatially calibrated and synchronized to ensure cross-modal alignment. Moreover, to embed an action quality task within gesture understanding, collected recordings are organized in classes of execution speed that mirror clinical evaluations of hand dexterity.
Baseline experiments highlight the dataset’s potential for gesture classification, gesture trigger detection, and action quality assessment. Thus, EHWGesture can serve as a comprehensive benchmark for advancing multimodal clinical gesture understanding.

\end{abstract}    

\section{Introduction}
\label{sec:intro}

Automatic hand gesture recognition is a complex yet significant challenge. Gestures are fundamental to human interaction and, along with facial expressions, constitute the non-verbal component of inter-subject communication \cite{gest_review2007}.

Computer systems capable of recognizing hand gestures can support various essential tasks. For instance, conventional applications include sign language recognition  \cite{sign_language}, augmented reality \cite{egogesture}, robotics, and human-computer interaction \cite{HaGRID}. Early gesture recognition methods primarily relied on wearable devices, such as gloves embedded with inertial measurement units \cite{saeed2022systematic}. Nowadays, computer vision methods, especially those based on deep learning \cite{dynamic_gesture_review2020}, have become state-of-the-art, offering a non-invasive and more natural interaction.

Gestures can be categorized as static or dynamic \cite{review_dynamic_2021}. Static hand poses are analyzed in still images, without any temporal evolution of the hand position. Large datasets, especially those based solely on RGB data, can be easily crowdsourced as they require minimal instrumentation (\eg, commercial webcams), as demonstrated by the HaGRID dataset by Kapitanov \etal \cite{HaGRID}. Also researchers in dynamic hand gesture understanding have leveraged crowdsourcing to collect large-scale datasets, as done by Materzynska \etal for the Jester dataset \cite{Jester}. However, this approach does not allow to capture multimodal data, despite the potential benefits of incorporating additional modalities such as depth or event-based data to handle the complex spatial and temporal variations typical of dynamic gestures. 

In particular, event-based gesture datasets that combine neuromorphic data with other modalities remain scarce, making cross-modality comparisons challenging. Furthermore, many dynamic gesture datasets rely solely on automated annotation using deep-learning-based hand tracking methods like MediaPipe \cite{HaGRIDv2}. While efficient, this may result in the absence of a strong ground truth, particularly for precise temporal segmentation of gestures. For example, tracking joint positions with a motion capture system would provide more accurate validation and enhance the reliability of gesture understanding benchmarks.

\begin{figure*}[!ht]
\centerline{\includegraphics[width=0.8\textwidth]{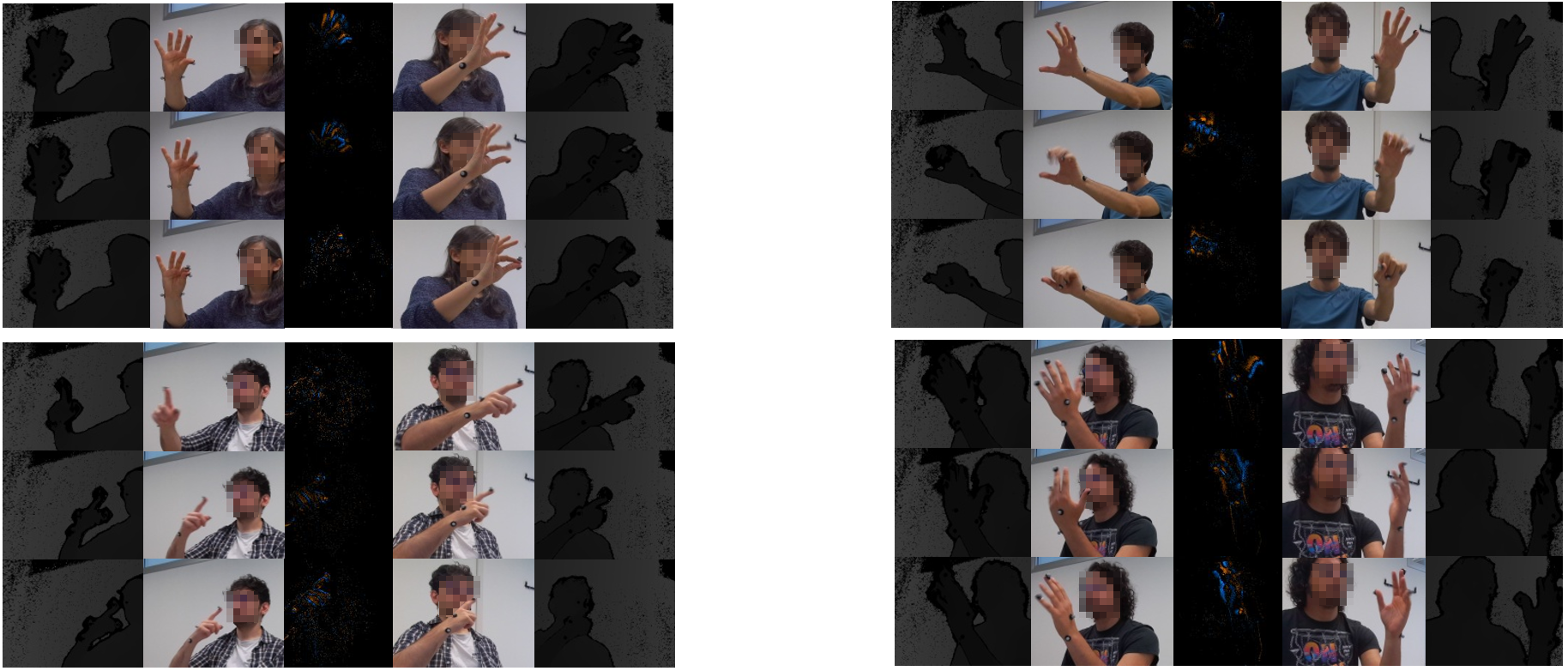}}
\caption{Dynamic gestures included in the dataset and keyframes acquired by the event (center) and RGB-D (left and right) cameras. From top-left: Finger tapping, hand opening and closing, finger-to-nose, and pronation-supination gestures.}
\label{fig:gestures}
\end{figure*}

Finally, dynamic gesture recognition can be extended to gesture \textit{understanding}, which involves not only classifying gestures within a given time window but also segmenting their \textit{triggering} phases (\eg, the tapping phase in finger pinching) \cite{HaGRIDv2} or assessing the quality of execution of the gesture (\eg, recognizing gesture speed) \cite{PECoP}. These tasks are central in innovative applications such as automatic hand dexterity assessment of clinical gestures. Indeed, a growing trend at the intersection of medicine and computer vision explores hand tracking and gesture recognition to automatically assess motor impairments caused by conditions such as Parkinson’s disease \cite{AMPRIMO2024102914}. This research typically involves multiple stages, including hand detection and tracking in video, gesture recognition to identify specific movements, and \gls{AQA} to quantify performance based on clinical ratings of disease severity \cite{PECoP}. However, large-scale, publicly available benchmarks in this domain are lacking \cite{AMPRIMO2024102914}. While pathological data are crucial for studying specific disease-related impairments, even extensive datasets from non-pathological subjects could improve the robustness of early-stage gesture recognition and tracking, which continue to face significant accuracy challenges.

This work introduces EHWGesture, a large-scale multimodal video dataset for dynamic gesture understanding. The dataset comprises five  gestures commonly used in clinical hand dexterity assessments, performed over 1,100 trials, each lasting 20s, for a total of more than six hours of recordings. Gestures were captured using two high-resolution RGB-Depth cameras positioned at orthogonal viewpoints, recording at 30 frame per seconds (fps), resulting in more than 2.6M RGB and depth frames. Additionally, a neuromorphic event camera sampled events at a high frequency (100 MHz). All recordings took place within the tracking volume of a motion capture system, which tracked key hand landmarks involved in the gestures. This \textit{gold standard} system provides precise ground truth labeling and enables validation against a highly accurate reference. All recording streams were synchronized to maintain temporal alignment across modalities. Spatial calibration data are also provided to facilitate the fusion of different data modalities from a spatial perspective. Finally, some gestures were performed at varying speeds, with volunteers following a metronome. This enables \gls{AQA} based on execution speed, adding a new dimension to gesture understanding that mimics clinical differences in hand dexterity. 

This work also presents a series of experiments using the newly collected dataset to establish baseline models for the main addressed tasks.  To summarize, its key contributions are:
\begin{itemize}
\item Providing a large-scale multimodal dataset, integrating synchronized and spatially calibrated RGB, depth, and event-based data.
\item Offering high-quality ground truth via a motion capture system for precise hand landmark tracking, enabling accurate validation and benchmarking.
\item Introducing controlled-speed recordings guided by a metronome, allowing for \gls{AQA} of the recorded gestures based on execution speed, similarly to clinical assessments.
\item Establishing baseline models on the EHWGesture dataset for gesture classification, temporal segmentation (\ie, triggering detection), and \gls{AQA}.
\item Investigating the impact of different modalities, input sequence length and frame rate on gesture understanding. 
\end{itemize}

\begin{table*}[t]
\centering
\small
\caption{Comparison of dynamic hand gesture datasets.  Acronyms: RGB – Red, Green, Blue, D - Depth, E - Event,
GR – Gesture Recognition,  
GD – Gesture Detection,  
GT – Gesture Triggering,  
AQA – Action Quality Assessment,  
HT – Hand Tracking, HGT - Hand Ground-Truth, SBJ -Subjects, PoV - Point of View, IR - Infra-Red}  

\label{tab:datasets}
\begin{tabularx}{\textwidth}{p{2.8cm}p{0.9cm}p{0.6cm}p{1.1cm}p{1.5cm}p{3.67cm}p{0.4cm}p{1.4cm}p{0.4cm}p{0.4cm}}
\toprule
\textbf{Dataset} & \textbf{Videos} & \textbf{SBJ} & \textbf{ Classes} & \textbf{Frames} & \textbf{Modality \& Resolution (px)} & \textbf{HGT} & \textbf{Tasks} & \textbf{PoV} & \textbf{Views} \\
\midrule
PD4T~\cite{PECoP} & 1,654 & 30 & 2 & - & RGB: 856$\times$480 & - & GR, AQA & 3rd & 1 \\
EgoGesture~\cite{egogesture} & 2,081 & 50 & 83 & 2,953,224 & RGB, D: 640$\times$480 & - & GR, GD & 1st & 1 \\
FHANDS~\cite{FHANDS} & 1,175 & 6 & 27 & 105,459 & RGB: 1920$\times$1080, D: 640$\times$480 & \checkmark & GR, HT & 1st & 1 \\
IPN Hand~\cite{IPNHand} & 200 & 50 & 13 & 800,000 & RGB: 640$\times$480 & - & GR, GD & 3rd & 1 \\
ChaLearnConGD~\cite{ChaLearn} & 22,535 & 21 & 249 & - & RGB, D: 240$\times$320 & - & GR, GD & 3rd & 1 \\
ChaLearnIsoGD~\cite{ChaLearn} & 47,933 & 21 & 249 & - & RGB, D: 240$\times$320 & - & GR & 3rd & 1 \\
Jester~\cite{Jester} & 148,092 & 1,376 & 27 & 5,331,312 & RGB: 100$\times$variable & - & GR & 3rd & 1 \\
NVGesture~\cite{NVGesture} & 1,532 & 20 & 25 & - & RGB, D, IR: 320$\times$240 & - & GR, GD & 3rd & 2 \\
DVSGesture~\cite{dvsgesture} & 1,342 & 29 & 11 & - & E: 128$\times$128 & - & GR, GD & 3rd & 1 \\
NavGesture~\cite{NavGesture} & 1,681 & 28 & 6 & - & E: 640$\times$480 & - & GR, AQA & 3rd & 1 \\
EB-HandGesture~\cite{EBHandGesture} & 9,000 & 5 & 6 & - & E: 640$\times$480 & - & GR, AQA & 3rd & 1 \\
\textbf{EHWGesture} & \textbf{3,300}$^\ast$ & \textbf{25} & \textbf{11} & \textbf{3,300,000}$^{\ast\ast}$ & \textbf{RGB: 1920$\times$1080, D: 640$\times$480, E: 320$\times$240} & \textbf{\checkmark} & \textbf{GR}, \textbf{GT}, \textbf{GD}, \textbf{AQA} & \textbf{3rd} & \textbf{3} \\
\bottomrule
\end{tabularx}
\vspace{2mm}
\raggedright
\small
$^\ast$ 1,100 recordings x (2 RGB-D cameras + 1 event camera) 
$^{\ast\ast}$ RGB + D + E frames, E accumulated using a 33 ms window
\end{table*}


\section{Related Works}
\label{sec:related_works}

Dynamic hand gesture datasets in the literature focus on different tasks depending on their applications (\ie, sign language \cite{bi2019graph}, conversational gestures, and interaction gestures). Large scale and open datasets including several clinical hand gestures are currently lacking, as most of the experiments rely on limited and private datasets \cite{AMPRIMO2024102914}. The PD4T  dataset \cite{PECoP} is the only exception, with data collected from 30 patients with Parkinson's disease to perform \gls{AQA}. Despite containing more then 1600 recordings, the dataset is limited to only two classes of clinical hand gestures (finger tapping, hand opening-closing) and one single modality (RGB). 

For a more comprehensive overview, this section broadens the perspective to gesture understanding for interaction, as these gestures are the most similar to those included in the EHWGesture dataset. Previous datasets aimed to address specific limitations in the field, such as small sample sizes—particularly in terms of subjects—, limited background diversity, restricted data modalities, and varying annotation types. Table \ref{tab:datasets} provides a structured comparison of the most relevant datasets in this domain, alongside the EHWGesture dataset.

Most datasets provide recordings from a third-person perspective, except for EgoGesture \cite{egogesture} and FHANDS \cite{FHANDS}, which are specifically designed for egocentric gesture recognition. All datasets except NVGesture and the proposed EHWGesture capture recordings from a single viewpoint \cite{NVGesture}. The largest dataset in terms of recorded frames and subjects, Jester \cite{Jester}, contains only RGB recordings, as it was obtained through crowdsourcing. EgoGesture, FHANDS, ChaLearn ConGD, and IsoGD \cite{ChaLearn} provide both RGB and depth data, whereas NVGesture also includes stereoscopic infrared recordings. For the latter, additional modalities, such as optical flow and segmentation masks, were derived from RGB, but Table \ref{tab:datasets} considers only raw acquired modalities.
Most datasets are annotated for gesture recognition using manually segmented gestures. However, IPN Hand \cite{IPNHand}, DVSGesture \cite{dvsgesture}, ChaLearn ConGD, and NVGesture also include annotations for continuous gesture detection, \ie, identifying gestures within an unsegmented recording that includes both gesture and non-gesture segments. FHANDS \cite{FHANDS} also provides ground-truth hand tracking data for key hand landmarks, obtained using a wearable system based on magnetic sensors. As a result, this dataset can also be used for hand tracking tasks in the context of hand-object interaction.
Datasets for event-based gesture recognition are less common, as neuromorphic cameras are still emerging. In event cameras, light reaching a pixel is converted into a voltage. Any deviation from a reference voltage is detected, and when it exceeds a predefined threshold, an event is triggered. These events are transmitted with extremely high temporal resolution (\eg, 100 MHz), and their data rate varies depending on the frequency of illumination changes in the scene. 

The three event-based datasets in Table \ref{tab:datasets} (DVSGesture \cite{dvsgesture}, EB-HandGesture \cite{EBHandGesture}, and NavGesture \cite{NavGesture}) do not provide complementary modalities for direct comparisons between neuromorphic vision and conventional approaches. DVSGesture was one of the first large-scale event-based gesture datasets and has been widely used for both isolated gesture recognition and continuous gesture detection \cite{dvsgesture,ghosh2019spatiotemporal,zhang2021event}. NavGesture examined gesture detection in static versus dynamic conditions, such as while walking, by combining an event camera with a smartphone. EB-HandGesture supports both gesture recognition and \gls{AQA} by recording gestures at three different speeds, though this aspect was not explored by the authors when introducing the dataset \cite{EBHandGesture}.

\noindent EHWGesture addresses current gaps by:
\begin{itemize} 
\item Providing the first dataset including multiple gesture from clinical assessment of hand dexterity, with timed gesture execution, mimicking real patients conditions, to incorporate \gls{AQA} into gesture recognition.
\item Including simultaneous and calibrated recordings from event, RGB and depth cameras from three different viewpoints, for cross-view and cross-modality comparisons
\item Offering ground-truth hand tracking through a high-precision motion capture system, ensuring fast and accurate temporal segmentation and spatial reference for key hand landmarks involved in the gesture.
\end{itemize}

The next section details the dataset and the collection pipeline used.


\section{EHWGesture Dataset}

EHWGesture was developed as a large-scale, multimodal dataset with high-quality ground truth annotations to support the recognition and characterization of five gestures commonly assessed in clinical evaluations. These gestures are based on the Unified Parkinson’s Disease Rating Scale (UPDRS), a widely used framework for evaluating hand dexterity in Parkinson’s disease \cite{UPDRS}. Four of these movements—finger tapping, hand opening and closing, hand pronation-supination, and finger-to-nose reaching— are dynamic, while one static gesture involves extending the arm forward to assess distal tremors.

The five gestures, captured across multiple modalities and viewpoints, are illustrated in Figure \ref{fig:gestures}.
Moreover, previous studies have shown that pre-training on other \gls{AQA} tasks, even from non-clinical domains, can aid Parkinson's video staging \cite{PECoP}. For this reason, the protocol was structured so that the finger tapping, pronation-supination, and opening-closing gestures were also performed at different execution speeds, embedding an \gls{AQA} task focused on pace recognition. Slowness in movement (\textit{bradykinesia}) is a hallmark of Parkinson’s disease. Thus, execution speed is typically linked with AQA for Parkinson's disease detection \cite{roalf2018finger}, further reinforcing the decision of embedding this kind of analysis in the dataset. This aspect could be leveraged by future clinical studies to support pretraining of deep learning models for clinical \gls{AQA} despite the current paucity of data from real patients.

 
\subsection{Dataset creation}

The dataset was created using video recordings from 25 healthy volunteers, all of whom provided written informed consent for data collection and sharing. Participants ranged in age from 24 to 65 years, with 7 female and 18 male subjects. 
The multimodal nature of the dataset imposed constraints on the sample size and recording conditions. Nonetheless, efforts were made to include participants with diverse hand shapes and skin tones.

\textbf{Instrumentation}: Recordings were conducted in a laboratory equipped with an Optitrack (OPT) system featuring six Prime13 cameras (1280×1024px resolution). OPT cameras operated at 120 Hz, covering a working volume of approximately 6×4×3 m\textsuperscript{3}. Passive reflective markers from the OPT system appeared in RGB frames and could also interfere with depth data by creating holes in the depth map. To minimize these effects, small reflective markers and a minimal marker schema were used, tracking only key hand landmarks relevant to each gesture rather than performing full hand tracking. 
Two RGB-Depth cameras (Microsoft Azure Kinect) were placed within the motion capture volume, positioned orthogonally so that their viewpoints intersected near the center of the capture area. Each Kinect device recorded RGB and depth frames at 30 fps, with resolutions of 1920×1080px and 640×480px, respectively. Between the two devices, an Inivation DVXplorer Lite event camera captured frontal recordings of the gestures, operating at 100 MHz with a resolution of 320×420px.

\textbf{Synchronization and calibration:} Temporal alignment of all recordings was achieved using an OptiTrack eSync2. The motion capture system coordinated all devices, managing acquisition start times and synchronizing exposure intervals for the Kinect cameras. The event camera operated independently, but a trigger event was generated each time an RGB-Depth frame was captured. This enabled the identification of events within the RGB frame acquisition, allowing event data to be accumulated into frames closely aligned with RGB images.
Each device underwent individual calibration using appropriate procedures. Additionally, stereo calibration was performed by capturing a checkerboard pattern displayed on a PC screen, with three OPT markers placed at specific corners. The PC screen was required as pixel refreshing generated clear events that were also detected by the DVXplorer Lite camera. Recordings were taken from 40 different positions within the tracking volume, corresponding to likely hand locations during gesture execution. This extrinsic calibration is included in the dataset to support cross-device alignment.

\textbf{Acquisition protocol:} Each subject sat comfortably in the center of the acquisition space. Full-body recordings were captured while subjects performed the gestures. Each subject executed the five tasks, first with one hand and then the other. Each task was continuously recorded for 20 seconds, mirroring clinical assessments where gestures are evaluated over time rather than in isolation. Moreover, longer videos can be segmented into different window sizes, introducing greater internal variability compared to isolated gestures. Finger tapping, hand opening and closing, and pronation-supination were timed using a metronome at three speeds: SLOW (75 bpm), NORMAL (115 bpm), and FAST (145 bpm). Subjects were instructed to follow the metronome as closely as possible. Executions excessively deviating from the request were discarded during acquisition and repeated again. Each task was recorded twice for data augmentation.

\textbf{Annotation}: Additional data were collected, including hand shape measurements (length and width) and illumination conditions, as recordings spanned multiple days. Each recording was labeled with the corresponding gesture name and, for timed gestures, the execution speed. Furthermore, motion capture data provided reference points for temporal segmentation, as continuous gesture repetitions generated periodic patterns in key hand landmarks. These patterns were segmented automatically or semi-automatically by identifying extreme points in the signals, allowing large datasets to be efficiently labeled with a strong ground truth.
To demonstrate the utility of motion capture data for this purpose, \textit{triggers} were automatically extracted, representing key moments within gestures. For example, in finger tapping, triggers corresponded to pinching instants; in hand opening and closing, the fully closed hand; in finger-to-nose reaching, the outward-reaching hand; and in pronation-supination, the moments when the palm or back of the hand faced the cameras.


\subsection{Dataset characteristics}

A total of 1,100 recordings were conducted (44 per subject, 22 per body side), resulting in 2,200 RGB-Depth videos and 1,100 event recordings. The synchronized RGB (1920×1080px) and depth (640×480px) frames across both cameras amount to 2,640,000 frames. While events are collected as sequential data on a spatial grid of 320×240px, aggregating them into frames at 30 fps (aligned with the RGB-D cameras) generates an additional 660,000 frames, bringing the total to 3,300,300 frames. These statistics refer only to raw data; additional derived modalities, such as 3D point clouds, optical flow, or segmentation masks, could be extracted from the depth and RGB data streams. 

The recorded hands were positioned between 30 and 100 cm from each camera, as subjects sat in a fixed location but adjusted their posture for comfort. Although gesture timing was enforced for some tasks, spatial amplitude was not restrained. Furthermore, the extended duration of each recording (20s) inherently introduced variability.

Considering the combined gesture recognition and \gls{AQA} tasks, recordings may be grouped in 11 classes, as reported in Table \ref{tab:datasets} (two free gestures plus three gestures $\times$ three velocities). Moreover, while we do not provide hand bounding boxes or full hand tracking models, this information can be easily obtained using tools such as \texttt{mediapipe-hands} \cite{Zhang2020MediaPipeHO}, as applied in our preprocessing (see supplementary materials). Additionally, motion capture data can be used to validate results from pre-trained 3D hand tracking networks, as demonstrated in \cite{AMPRIMO2024106508}. Finally, given the available depth data and motion capture annotations, the dataset could also support the development of new multimodal 3D hand tracking models, particularly in pre-training phases.


\section{Experiments}

We showcase the potential of the EHWGesture dataset for gesture understanding through two experimental use cases and their respective baseline models.

First, we train models for gesture classification and \gls{AQA}. These tasks are key to advancing gesture recognition systems for clinical assessment and quality-based analysis. This experiments employ unsegmented recording windows, making it more challenging than isolated gesture recognition, as windows may contain incomplete gestures. For gesture recognition, the five gestures, ignoring velocity classes, are classified. For \gls{AQA}, the objective is to classify gestures based on the three execution speed classes (SLOW, NORMAL, FAST). 

The second experiment focuses on detecting trigger events within recordings, using motion capture data as ground truth. This demonstrates the applicability of EHWGesture also to other field than clinical assessment, such interactive gesture research, and highlights the advantages of motion capture for precise annotation.

The data and the code used for these experiments are available at \href{https://github.com/smilies-polito/EHWGesture}{https://github.com/smilies-polito/EHWGesture}.



\subsection{Architectures}

The dataset was benchmarked using three different convolutional architectures: PhiNet-3D \cite{PaissanAncilotto2022Phinets}, 3D ResNet-50 \cite{He2015DeepRL}, and 3D ResNeXt-152 \cite{xie2017aggregated}. The 3D architectures were derived from their respective 2D versions following the approach in \cite{kopuklu2019resource}. These models cover a broad range of computational complexities and network sizes, from the smallest PhiNet with $4.9M$ parameters to the largest ResNeXt with $116M$ parameters.

The two addressed tasks—gesture recognition and \gls{AQA}—require different types of information. Gesture recognition primarily relies on spatial features, as shown in previous studies that solved the task using single-frame models \cite{HaGRID}. In contrast, \gls{AQA} depends on temporal information, requiring models that effectively capture motion dynamics \cite{NavGesture}.

\begin{figure}[b]
  \centering\includegraphics[width=0.58\linewidth]{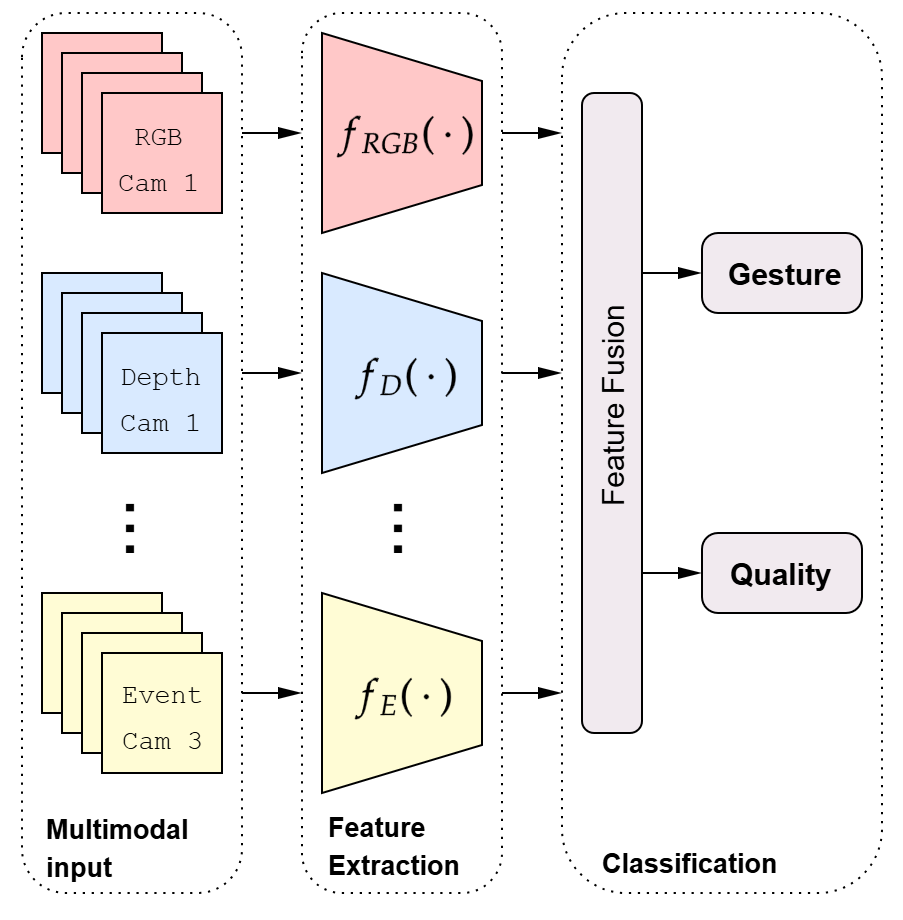}

   \caption{Proposed architecture for multimodal video analysis}
   \label{fig:multimodal_architecture}
\end{figure}

To process the multimodal data, the networks were used to extract features from different input streams. A late feature fusion strategy, similar to \cite{carreira2017quo}, was then applied to classify the multimodal input, as illustrated in Figure \ref{fig:multimodal_architecture}. This approach enables the evaluation of how integrating RGB, depth, and event-based representations separately contributes to the gesture understanding task.

\subsection{Training Setup}

The training data consists of cropped RGB, depth, and event-based frames processed as described in the supplementary materials. All input frames were resized to a uniform resolution of 240p to ensure consistency across modalities.
We defined a train/validation split that reserves data from five subjects for evaluation.
Networks were trained using the \texttt{SGD} optimizer with an initial learning rate of $10^{-3}$, which was reduced by a factor of $0.1$ every $10$ training epochs. 

\textbf{Multimodal Contrastive Pretraining:}
Before fine-tuning for classification, each network underwent a multimodal contrastive pretraining phase. The SimCLR \cite{chen2020simple} framework was adapted for contrastive pretraining with multimodal inputs. In particular, to accommodate the dataset’s multi-camera, multimodal nature, the original data augmentation strategy was modified. Specifically, given a sequence of frames $x_t^{(c,m)}$ at timestamp $t$, with modality $m$ and camera $c$, positive and negative pairs were defined as follows:

\begin{itemize}
\item Positive pairs: frame sequences with the same timestamp but from different cameras or modalities:
\begin{equation}
\begin{split}
    &P: (x_t{_1}^{(c_1,m_1)}; x_t{_2}^{(c_2,m_2)}) \text{ with } t_1=t_2 \\
    & \text{ and either } c_1 \neq c_2 \text{ or } m_1 \neq m_2.
\end{split}
\end{equation}
\item Negative pairs: frame sequences starting from different timestamps or referring to different subjects
\begin{equation}
\begin{split}
    &N: (x_t{_1}^{(c_1,m_1)}; x_t{_2}^{(c_2,m_2)}) \text{ with } t_1\neq t_2 
\end{split}
\end{equation}
\end{itemize} 
The goal of this pretraining was to encourage the network to learn modality-invariant and viewpoint-invariant representations that can be exploited for downstream tasks. Further pretraining considerations are reported in the supplementary materials.


\begin{figure*}[]
    \centering
    \begin{subfigure}{0.46\linewidth}
        \centering
        \includegraphics[width=.85\linewidth]{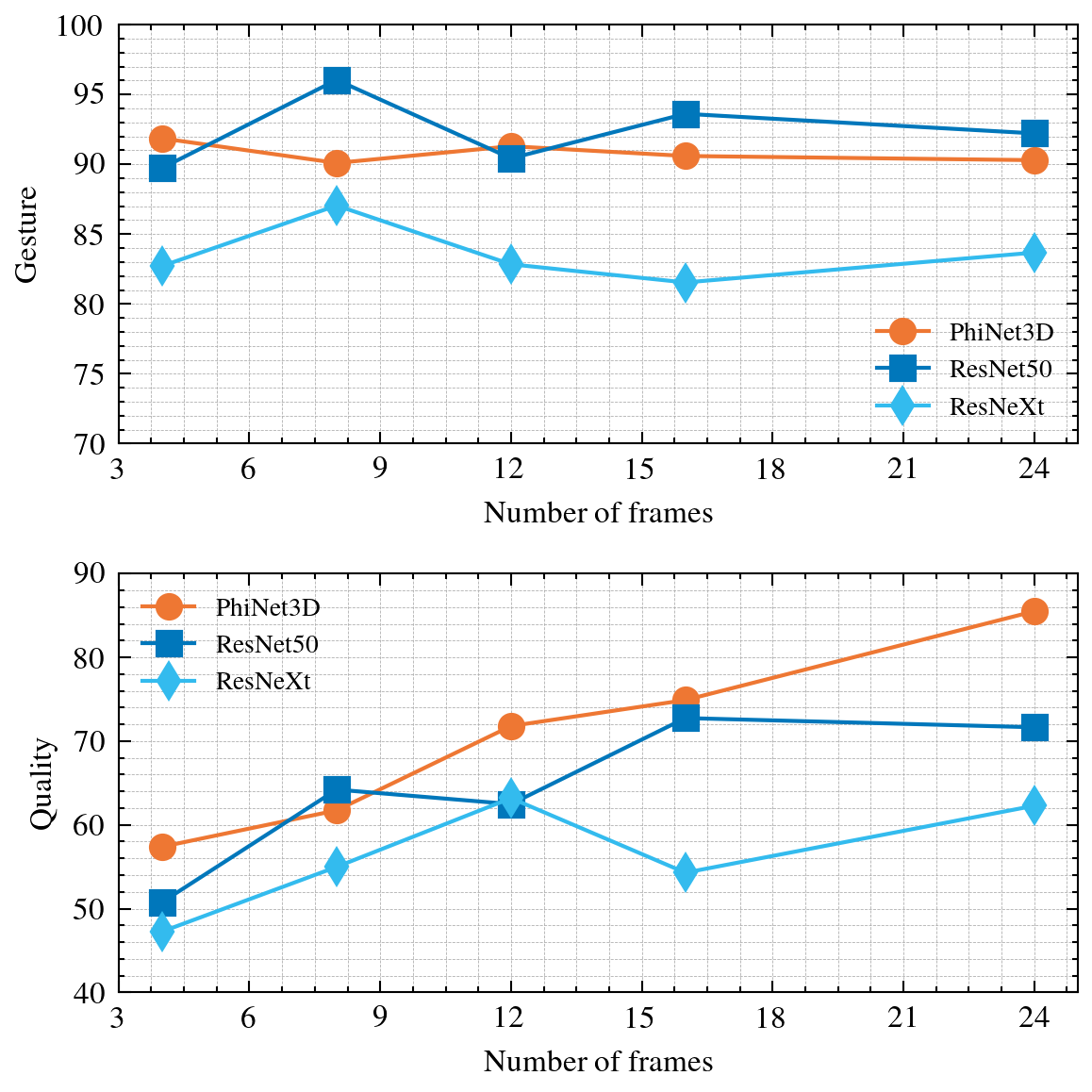}
        \caption{Impact of the number of window sizes on network accuracy.}
        \label{fig:input_length}
    \end{subfigure}
    \hfill
    \begin{subfigure}{0.46\linewidth}
        \centering
        \includegraphics[width=.85\linewidth]{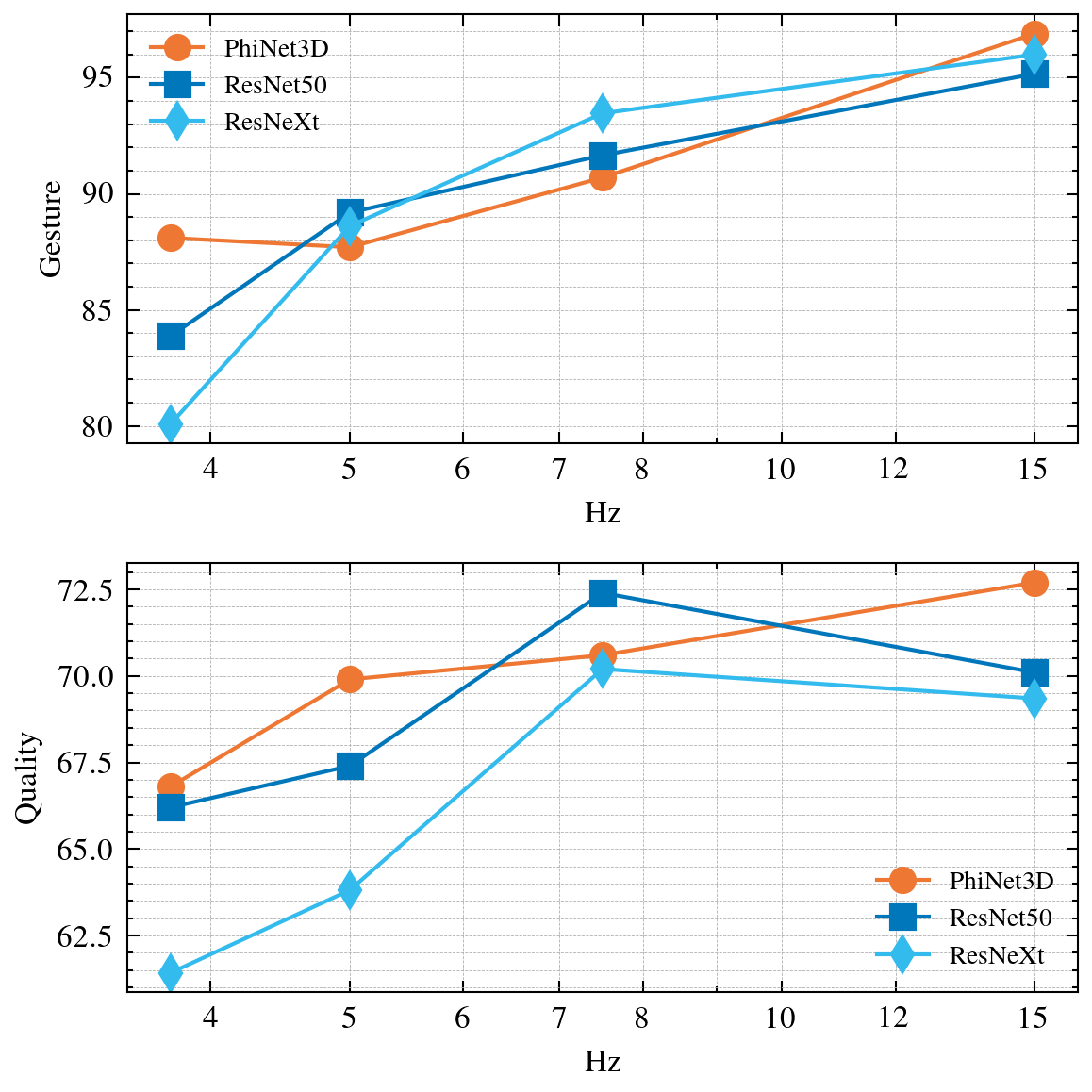}
        \caption{Impact of input framerate on network accuracy.}
        \label{fig:fps}
    \end{subfigure}
    \caption{Comparison of the impact of input length and framerate on network accuracy for the two tasks.  (a) Quality estimation shows a strong dependency on window length, while gesture classification is almost independent. (b) Higher framerates are beneficial for gesture classification, although leading to increased computational complexity. Quality estimation benefits slightly from temporal downsampling.}
    \label{fig:input_analysis}
\end{figure*}

\subsection{Sequence Length and Framerate Impact}
\label{sec:fps_and_window_len}

Given the different temporal dependencies of the two proposed tasks, we analyzed the impact of frame rate and time window length on the classification performance of the various networks.
Figure \ref{fig:input_length} illustrates how input length affects network accuracy. The results indicate that gesture recognition is largely time-invariant, with accuracy remaining stable across windows of different lengths. In contrast, \gls{AQA} benefits significantly from longer temporal contexts, as extended sequences provide more information on execution speed and consistency.
Figure \ref{fig:fps} presents the effect of temporal subsampling on classification accuracy. While lower frame rates allow for the analysis of longer time windows, they appear to degrade performance for gesture recognition. This suggests that key temporal features are already captured at higher sampling rates, and further extending the time window through undersampling does not add meaningful discriminative information. In contrast, for quality classification, ResNet and ResNeXt show a slight performance improvement when data is downsampled to $7.5$ fps.


\subsection{Trigger detection}

For trigger detection, we used a baseline pipeline that leverages hand tracking via \texttt{mediapipe-hands}. The processing for each recording consisted of:

\begin{enumerate}
\item Extracting hand landmarks using \texttt{mediapipe-hands} from RGB data, analyzing the \texttt{Main} and \texttt{Sub} camera separately.
\item Computing reference gesture trajectories for each gesture.
\item Merging reference trajectories from both views using the arithmetic mean and applying a 1D convolution with a fixed-size smoothing window to refine the results.
\item Identifying local extrema in the resulting trajectory, which correspond to the triggers of interest.
\end{enumerate}

For finger tapping, we used the distance between the thumb tip and index tip; for hand opening and closing, the distance between the middle finger and wrist; for the finger-to-nose gesture, the trajectory of the index tip across the image; and for pronation-supination, the distance between the index tip and pinkie tip. We experimented with different smoothing window sizes (3, 5, 7) to assess variations across execution speeds.

To evaluate baseline performance, we used the mean absolute error (MAE) to measure temporal delay, detection accuracy to quantify the number of correctly identified triggers compared to ground truth, and the false detection ratio (FDR) to account for spurious triggers generated by the detection pipeline.

\section{Results}


\subsection{Gesture classification and action quality}

\begin{figure*}[]
\centering\includegraphics[width=.90\linewidth]{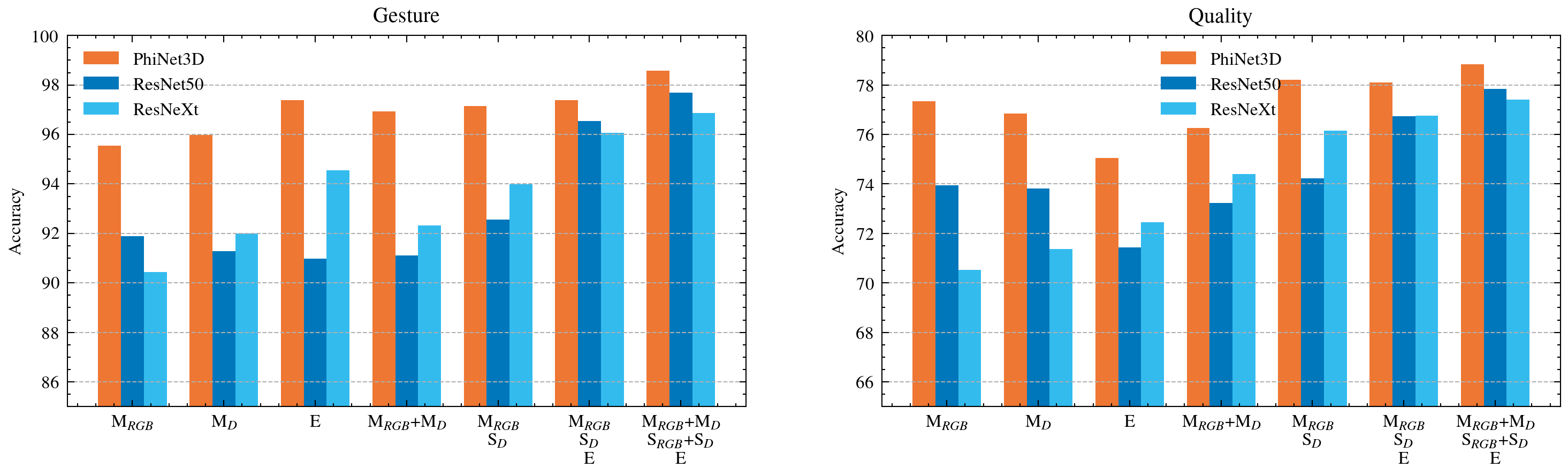}

\caption{Performance of the tested architectures across both tasks. Increasing the number of input modalities generally improves performance, particularly for larger networks. M: main camera, S: sub camera, E: event, D: depth}
\label{fig:ablation}
\end{figure*}

To assess the impact of multimodal inputs on performance, the different architectures were trained on both tasks using an increasing number of input modalities. The results, summarized in Figure \ref{fig:ablation}, illustrate performance trends across different architectures and input configurations.

Interestingly, model performance does not show a strong correlation with the complexity or size of the feature extractors used; even relatively lightweight architectures achieve competitive classification results in both tasks. Across all architectures, unimodal inputs yield similar performance, with depth information providing a slight improvement over RGB data from the same camera. Likewise, event-based data perform comparably to other single-modality inputs.

Fusing RGB and depth from a single camera results in a modest performance boost ($+0.1\%$ for gesture recognition and $+1.6\%$ for quality estimation on average). In contrast, incorporating data from two distinct cameras leads to a more substantial improvement ($+1.2\%$ for gesture recognition and $+3.2\%$ for quality estimation on average). Ultimately, leveraging all three modalities in combination yields the highest performance gains across both tasks, with an average accuracy increase of $+3.3\%$ for gesture detection and $+4.5\%$ for quality estimation.

\subsection{Trigger detection}

Results for trigger detection on the five validation subjects are presented in Table \ref{tab:task_stats}. For each task, we report only the result achieved using the optimal smoothing window (according to accuracy). The findings indicate that detecting gesture triggers is generally straightforward, with high detection accuracy across all gestures, despite the simplicity of the baseline approach. However, the main challenge lies in precisely identifying the exact timing of the trigger event, as both MAE and FDR exhibit high values with large standard deviations. Consequently, EHWGesture may serve as a  benchmark for trigger detection, particularly for gestures such as pronation-supination and finger-to-nose reaching.

\begin{figure}[b]
  \centering
  \includegraphics[width=0.80\linewidth]{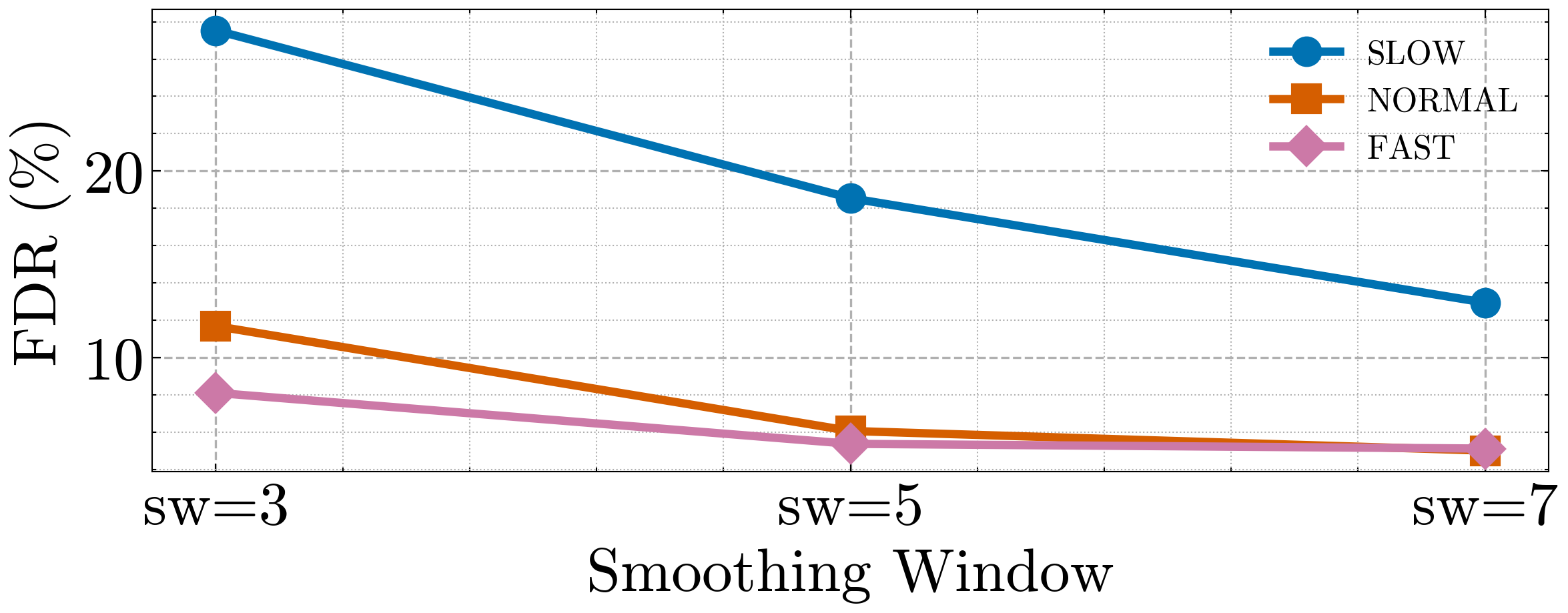}

   \caption{Percentage of wrong trigger detections over total detections considering different smoothing windows. Slower movements improve the most from longer windows.}
   \label{fig:smoothingvsspeed}
\end{figure}

We also examine the relationship between execution speed and smoothing window size. As shown in Figure \ref{fig:smoothingvsspeed}, trigger detection for slow movements significantly benefits from longer smoothing windows, which help reduce false detections. In contrast, FAST and NORMAL gestures achieve comparable performance with smoothing windows of size 5 or 7. This occurs because slower gestures are more ambiguous to segment, as their minima points often correspond to extended plateaus where multiple consecutive frames may be misinterpreted as trigger events. Therefore, integrating \gls{AQA} predictions related to execution speed into this baseline could enable automatic tuning of smoothing window sizes for the triggering task.

\begin{table}[ht]
\centering
\caption{Trigger Detection performance (Best Smoothing Window \textit{sw}).   Tasks: FT-finger taping, OC-hand opening-closing, PS - Pronation-Supination, NOSE-finger-to-nose.}

\label{tab:task_stats}
\small
\centering
\begin{tabular}{p{2cm}p{1.25cm}ccc}
\hline
Task & MAE (s) & Acc. (\%) & FDR (\%) \\
\hline
FT (sw=3)          & 0.12 & 97.37 $\pm$ 3.54 & 16.11 $\pm$ 15.13 \\
OC (sw=3)           & 0.11 & 98.80 $\pm$ 1.78 & 20.62 $\pm$ 18.87 \\
PS  (sw=5)         & 0.27 & 97.26 $\pm$ 5.16 & 7.98 $\pm$ 8.70  \\
NOSE (sw=3)         & 0.48 & 98.07 $\pm$ 3.08 & 28.97 $\pm$ 19.52 \\
\hline
\end{tabular}
\end{table}

\section{Discussion} 

The results obtained in the experiments demonstrate that EHWGesture can serve as a benchmark for multimodal clinical gesture understanding, supporting diverse and relevant tasks such as gesture classification, \gls{AQA}, and trigger detection. These two tasks in particular still offer margin for improvement and are pivotal for supporting the creation of more advanced models for \gls{AQA} to use with real pathological data, as done by Dadashzadeh \etal in \cite{PECoP}. Multimodality, the main contribution of this work, proved essential for improving classification models, and its impact goes beyond the specific gestures of this dataset. EHWGesture can facilitate further research on combining either the three provided raw modalities or additional derived modalities. Moreover, the two Kinect devices and the event camera captured gestures from three distinct perspectives, influencing model training. This aspect opens possibilities for exploring the impact of different viewpoints on gesture understanding.

\textbf{Ethical considerations}:
Data collection adhered to GDPR, and all subjects provided written consent to share their data for non-commercial, research-only purposes. Videos are publicly released in an anonymized format, with faces blurred in all frames. We used SAM2 to detect overlapping regions of hands and faces, applying blurring only to the non-intersecting portion. This approach ensures that identifying information cannot be misused for purposes such as generating deepfakes, or facilitating identity theft. Subjects’ identities were not stored, as each participant was assigned an anonymous identifier during data collection.

\textbf{Biases}:
As this dataset prioritizes providing a comprehensive multimodal data source, subject diversity was somewhat limited. The small number of volunteers does not currently allow for extensive stratification, with only five subjects not conforming to the Caucasian phenotype. However, we plan to extend the dataset to improve subject diversity in future iterations, possibly including also  pathological subjects. 

\textbf{Limitations:} 
The use of a motion capture system implied recording all trials in the same environment. This limits background diversity in RGB frames, though other modalities remain unaffected. This limitation is, however, coherent with clinical gesture assessment, since examinations are often conducted in standardized settings. In addition, event frames may be influenced by different lighting conditions; while illumination levels were annotated during recordings, extreme lighting conditions were not tested, as they could severely impact reliability of motion capture data. As previously noted, the dataset is biased toward fair skin tones, which may affect the robustness of models trained solely on the RGB modality. However, the multimodal nature of the dataset may help mitigate this limitation.
Finally, the baseline trigger detection method does not incorporate a training phase but instead relies on pretrained models and deterministic signal processing. As a result, it may be fragile and prone to inconsistencies when applied to real-world scenarios. Nonetheless, we observed that its performance could be improved by integrating \gls{AQA} information estimated by the trained multimodal models.

\section{Conclusion}
This work introduced EHWGesture, a large-scale benchmark for multimodal  gesture understanding. The dataset includes gestures from clinical hand dexterity assessment and may support the development of automated models aimed at this application. Additionally, EHWGesture is the first gesture dataset to simultaneously integrate RGB, depth, and event data captured from three different viewpoints while also incorporating an \gls{AQA} based on gesture execution speed. Baseline experiments demonstrated the dataset's potential and provided insights for future research using this resource.

\section*{Acknowledgments}
This study is supported by SERICS project (PE00000014) under the MUR National Recovery and Resilience Plan funded by the European Union - NextGenerationEU.
{
    \small
    \bibliographystyle{ieeenat_fullname}
    \bibliography{main}
}

\end{document}